\documentclass[10pt,twocolumn,letterpaper]{article}

\usepackage[pagenumbers]{cvpr} 

\usepackage{graphicx}
\usepackage{amsmath}
\usepackage{amssymb}
\usepackage{booktabs}

\usepackage[pagebackref,breaklinks,colorlinks]{hyperref}

\usepackage[capitalize]{cleveref}
\crefname{section}{Sec.}{Secs.}
\Crefname{section}{Section}{Sections}
\Crefname{table}{Table}{Tables}
\crefname{table}{Tab.}{Tabs.}

\def\cvprPaperID{5411} 
\def\confName{CVPR}
\def\confYear{2023}

\usepackage{import}

\usepackage{xcolor}
\definecolor{purple}{cmyk}{0.44, 0.92, 0, 0.01}

\definecolor{gold}{rgb}{0.828, 0.683, 0.214}

\newcommand{\R}{\mathbb{R}}
\newcommand{\vc}[1]{\mathbf{#1}}

\def\fg#1{Fig.~\ref{fig:#1}}
\def\sec#1{Section~\ref{sec:#1}}
\newcommand{\eq}[1] {Eq.~\ref{eq:#1}}
\newcommand{\tab}[1] {Table~\ref{tab:#1}}

\begin{document}

\title{Synthesizing Photorealistic Virtual Humans Through Cross-modal Disentanglement}

\author{Siddarth Ravichandran, \hspace{0.12cm} Ond\v{r}ej Texler, \hspace{0.12cm} Dimitar Dinev, \hspace{0.12cm} Hyun Jae Kang  \\
	NEON, Samsung Research America \\
	{\tt\small \{siddarth.r, o.texler, dimitar.d, hyunjae.k\}@samsung.com}
}


\twocolumn[{%
	\renewcommand\twocolumn[1][]{#1}%
	\maketitle
	\begin{center}
		\centering
		\captionsetup{type=figure}
		\def\svgwidth{\hsize}
\begingroup%
  \makeatletter%
  \providecommand\color[2][]{%
    \errmessage{(Inkscape) Color is used for the text in Inkscape, but the package 'color.sty' is not loaded}%
    \renewcommand\color[2][]{}%
  }%
  \providecommand\transparent[1]{%
    \errmessage{(Inkscape) Transparency is used (non-zero) for the text in Inkscape, but the package 'transparent.sty' is not loaded}%
    \renewcommand\transparent[1]{}%
  }%
  \providecommand\rotatebox[2]{#2}%
  \newcommand*\fsize{\dimexpr\f@size pt\relax}%
  \newcommand*\lineheight[1]{\fontsize{\fsize}{#1\fsize}\selectfont}%
  \ifx\svgwidth\undefined%
    \setlength{\unitlength}{1144.5bp}%
    \ifx\svgscale\undefined%
      \relax%
    \else%
      \setlength{\unitlength}{\unitlength * \real{\svgscale}}%
    \fi%
  \else%
    \setlength{\unitlength}{\svgwidth}%
  \fi%
  \global\let\svgwidth\undefined%
  \global\let\svgscale\undefined%
  \makeatother%
  \begin{picture}(1,0.22114273)%
    \lineheight{1}%
    \setlength\tabcolsep{0pt}%
    \put(0,0){\includegraphics[width=\unitlength,page=1]{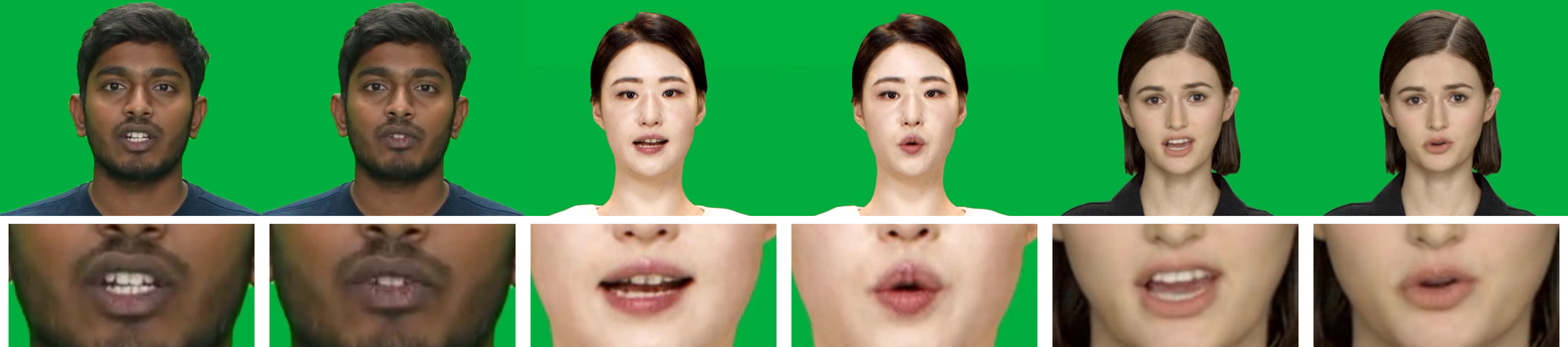}}%
  \end{picture}%
\endgroup%
\caption{
			Three sequences of talking faces generated using the proposed framework. Our 
			method produces high-texture quality, sharp images, and correct lips shapes 
			that are inline with the spoken audio. See the zoom-in patches, our method 
			faithfully reproduces fine textural and content details such as teeth.}\label{fig:teaser}
	\end{center}
}]


\begin{abstract}
Over the last few decades, many aspects of human life have been enhanced with
virtual domains, from the advent of digital assistants such as Amazon's Alexa
and Apple's Siri to the latest metaverse efforts of the rebranded Meta. These 
trends underscore the importance of generating photorealistic visual depictions 
of humans.
This has led to the rapid growth of so-called deepfake and talking-head generation 
methods in recent years. Despite their impressive results and popularity, they usually 
lack certain qualitative aspects such as texture quality, lips synchronization, or 
resolution, and practical aspects such as the ability to run in real-time.
To allow for virtual human avatars to be used in practical scenarios, we 
propose an end-to-end framework for synthesizing high-quality 
virtual human faces capable of speaking with accurate lip motion 
with a special emphasis on performance.
We introduce a novel network utilizing visemes as an intermediate audio
representation and a novel data augmentation strategy employing a hierarchical
image synthesis approach that allows disentanglement of the different modalities 
used to control the global head motion. Our method runs in real-time,
and is able to deliver superior results compared to the current state-of-the-art.
\end{abstract}


\section{Introduction}
\label{sec:intro}
With the metaverse gaining traction in the media and the rapid 
{\em virtualization} of many aspects of life, virtual humans 
are becoming increasingly popular. Propped up by technological advances in
hardware, GPU computing, and deep learning, digitization of humans
has become a very active research topic.

One challenging problem is creating virtual avatars that attempt to
faithfully recreate a specific human indistinguishable from their real 
counterparts. This
entails a perfect visual recreation, with high quality textures that
recreate the minutia that make a human visually realistic, e.g., hair
that appears to be composed of individual strands, clothes that appear
made out of fabric, and pores or vellus hair on the skin. Recent advances
in neural rendering have achieved considerable success in creating very
convincing images that are able to visually recreate such fine details.
Additionally, many use-cases involving virtual humans require them to appear on
large screens, e.g., human-size displays in airport halls, hotels, and billboards.
For such use-cases, it is vital for the virtual humans to be rendered in a
relatively high resolution.

Unfortunately, rendering quality alone is not sufficient to recreate a
convincing virtual human that can be interacted with. Just as important
as the visual fidelity is the fidelity of the motion. Specifically, for
speech-driven avatars, the way the avatar talks, opens its mouth, and moves
its lips is crucial. The human perception system is very sensitive to even
the slightest mistake when it comes to the face, specifically the appearance
of mouth and lips during speech. In order for a virtual avatar to be believable,
its mouth and lip motion must match the speech a human listener hears. 

In this work, we introduce a novel training regime that utilizes 1-dimensional 
audio features such as visemes~\cite{fisher1968confusions} or wav2vec~2.0~\cite{wav2vec2}. 
These are higher-level 
features compared to traditional features such as MFCC~\cite{mermelstein1976distance} 
which require more complex encoders. 
This enables us to efficiently synthesize talking heads with high-quality rendering and lip
motion. However, this alone is insufficient for creating an avatar suitable for the
aforementioned use-cases. In order to have more control over the avatar, it needs to
be conditioned on multiple constraints in addition to speech, such as head pose or body
shape. With speech represented as a 1-dimensional vector of visemes, head pose represented
as rotation angles, and body shape represented as a 2-dimensional outline, we
quickly run into a problem with different modalities among our data representation.

Training deep neural networks on multimodal data is a difficult problem,
especially when the different modalities are innately correlated. The network then 
tends to overfit on the modality that is easier to learn from, and this kind of 
overfitting leads to very poor performance at testing time. While there exist 
many techniques to mitigate traditional overfitting such as reducing number 
of parameters, regularization, etc., they are less effective when the overfitting 
is caused by multiple modalities. To break the correlations between these modalities,
we introduce a novel data augmentation strategy: for every given lip shape, we use 
a generative oracle network to synthesize photorealistic images of the same lip 
shapes in a variety of head poses. This requires a generative network capable of
producing images that are as high-resolution as our training data to preserve sharpness
in important areas such as teeth.

Even with modern hardware and the latest research,
generative neural networks still struggle with generating high-resolution images,
and the resolution usually comes at the cost of quality. When increasing the resolution
and the complexity of what the network is asked to generate, the generative network 
often loses the semantic quality or other qualitative aspects of the result 
images. Increasing the network's capacity alone by adding more learnable
parameters does not necessarily help and can introduce noise. To alleviate
this, we propose a hierarchical approach to generating a high-quality synthetic
image of a talking face that preserves the quality in the mouth region via
high-resolution supervision.

In order to reproduce a believable interaction with a virtual avatar, inference speed
is paramount. Our paper presents an efficient framework for creating high-quality virtual 
artificial humans in real-time where each of the aforementioned concerns is addressed. 
The contribution of our works is:

\begin{itemize} 
	\item A data augmentation method to disentangle audio and visual modalities so 
	that the whole framework can be trained end-to-end.
	\item A hierarchical {\em outpainting} approach which allows for generation
	of high-resolution synthetic data.
	\item An end-to-end framework that utilizes 2-encoder-2-decoder neural network
	architecture and leverages synthetic data.
\end{itemize}

\section{Related Work}
\label{sec:related}

\begin{figure*}[!ht]
 \includegraphics[width=\textwidth]{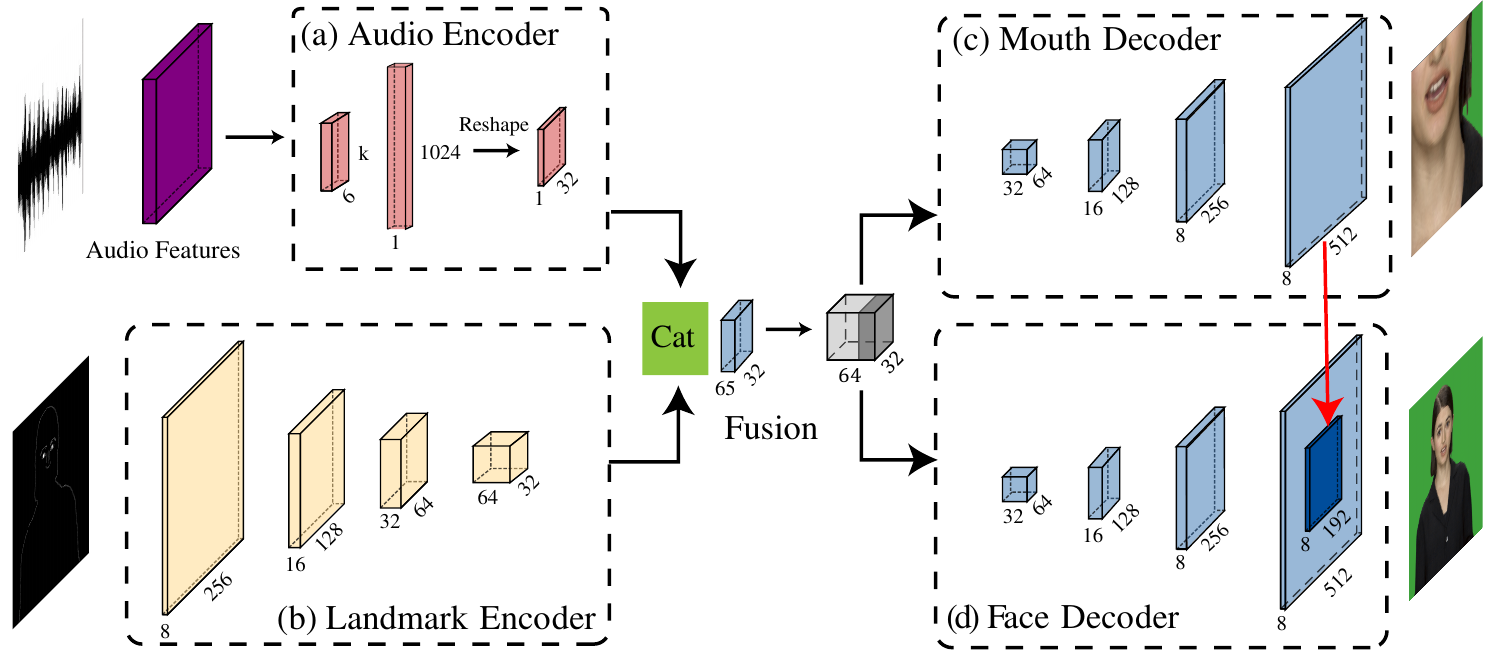}
\caption{
Overview of our multi-modal neural rendering pipeline. We generate $k$ audio features
of the provided audio, and sample a sequence of keypoint drawings and 
contours which are then mapped to a shared latent space using Audio Encoder~(a) 
and the Landmark Encoder~(b) respectively; which is then passed through the 
mouth~(c) and face~(d) decoders to produce the mouth and face images respectively. 
The mouth feature maps are pasted into the head feature map as denoted by the 
red arrow. During inference, only the output of the head decoder is used.
}
\vspace{-0.10in}
\label{fig:Inference}
\end{figure*}

{\bf Talking faces.}
In the last several decades, many methods to generate videos of talking faces 
have been developed, and most of the recent ones heavily leverage generative 
adversarial networks--GANs~\cite{Goodfellow14}. Having the photorealism as a main 
objective, various methods, colloquially known as 
``deepfakes''~\cite{Suwajanakorn2017, Vougioukas19, Thies2016face2face, Bregler1997, Khakhulin2022ROME, Zakharov2019FewShotAL, Drobyshev2022_megaportrait}, 
have been proposed and successfully used to synthesize images of human faces.
Talking face animation can be categorized into face re-enactment approaches~\cite{hong2022depth, wang2021facevid2vid, wayne2018reenactgan, OneShotFace2019, nirkin2019fsgan, nirkin2022fsganv2}
and speech driven approaches~\cite{song2018talking, zhou2019talking, zhang2022text2video, Yang2020_MakeItTalk, Richard_2021_ICCV, Chen2019_hierarchical, Prajwal2020lip}.
Speech driven models have used a variety of
intermediate representations such as 2D keypoints~\cite{Suwajanakorn2017, Chen2019_hierarchical}, 
meshes~\cite{Cudeiro2019capture,Thies2020neural, Richard_2021_ICCV}, which themselves can be 
derived from various audio features such as MFCC, spectral coefficients, or 
DeepSpeech features~\cite{Hannun2014deep}. Some methods directly produce
 talking faces from audio features without additional geometric representations ~\cite{zhang2022text2video} and some ~\cite{zhou2019talking, Prajwal2020lip} use an additional identity image or video as guiding signal.
A common issue these methods face is the reduced high-frequency motion,
such as lip motion during speech. Unrealistic lip movement is easy to notice, 
and disrupts the overall experience.
This is partially addressed in the work of {\em Prajwal et al.}~\cite{Prajwal2020lip},
they utilize SyncNet~\cite{Chung2016OutOT} to supervise the audio and lips synchronization,
which results in a dynamic mouth movement. However, they do not consider 
the photorealistic appearance. 
As the mandate of our work is to create virtual humans that are indistinguishable from
real people, we need to secure both realistic lip motion and photorealistic textures.

{\bf Hierarchical image generation.}
Synthesizing high-resolution images while retaining various fine details and 
overall sharpness is a difficult task. Researchers have been approaching 
this problem from various angles, and one of widely used solutions is to tackle 
this challenge in a hierarchical fashion.
InsetGAN~\cite{Fruhstuck2022insetgan} propose to combine multiple pre-trained
GANs, where one network generates the coarse body image, and other networks refine
certain regions.
The work of {\em Li et al.}~\cite{Li2021_collaging} proposes to have several 
class-specific generators that are used to refine certain semantic regions 
of the previously synthesized image.
TileGAN~\cite{Fruehstueck2019_TileGAN} presents a way to combine small-resolution
synthesized images in order to produce a single high-resolution texture.
{\em Xu and Xu}~\cite{Xu_Xu_2022} hierarchically compose the image from smaller patches, 
and {\em Iizuka et al.}~\cite{Iizuka2017} solve inpainting task using two discriminators, 
one for the whole image, and the other for a small area centered at the completed region.
Although those approaches are not directly applicable in our talking virtual
human scenario, our pipeline also leverages the hierarchical composition.

{\bf Multi-modal training.} To puppet a virtual human, multiple different 
inputs are usually needed, e.g., speech to control the lips and head-pose 
to specify head position and rotation. Those inputs--modalities 
are usually strongly correlated, which translates in the need of a multi-modal
training~\cite{talkingheadmultimodal,zhou2021pose,Zhang_2021_CVPR, Vougioukas2020}.
NeuralVoicePuppetry~\cite{Thies2020neural} trains a generalized expression 
network and an identity targeted neural renderer method decoupling the 
modalities using an intermediate mesh representation. 
MakeItTalk~\cite{Yang2020_MakeItTalk} propose to disentangle the modalities by learning 
displacements from a reference facial landmark using a separate module and 
perform rendering from the intermediate 2-dimensional landmark representations. 
Similarly, {\em L. Chen et al.}~\cite{Chen2019_hierarchical} use a learned 
high-level hierarchical structure to bridge audio signal with the pixel image.
MeshTalk~\cite{Richard_2021_ICCV} deals with the modality disentanglement in a
lower dimensional space by employing the cross-modality loss.
Many of these methods require an intermediate representation 
for the face in the form of drawings or mesh, hindering their
ability to produce diverse lip shapes. We address these limitations in our work
by using a mix of 1-dimensional and 2-dimensional representations, requiring
modality disentanglement in order to synthesize speech-driven talking faces.

\section{Our Approach}
\label{sec:method}
We present a framework to generate photorealistic virtual human avatars that 
can be streamed in real-time to enable interactive applications. We train our 
system on a single identity and during the inference we drive it using
audio features and partial face keypoints, as shown in~\fg{Inference}.
Due to bias in the way humans speak, there is a strong correlation between head
motion and speech which causes the network to model undesirable relations. 
To combat that, we propose a novel data augmentation strategy employing keypoint
mashing and a high-resolution generative oracle network to achieve disentanglement of
lip motion from upper face motion.
This allows synthesizing videos with arbitrary upper face keypoints and audio.

\subsection{Input Features}\label{sec:audio2lips}

{\bf Audio features.}
One of the crucial requirements for our method is the ability to faithfully 
reproduce motion and shapes of mouth and lips. Thus, it is vital to have an 
intermediate audio representation that is capable of capturing those geometric
characteristics. Visemes~\cite{fisher1968confusions} are lip shape categories
that describe the basic lip shapes used to make sounds common in human speech.
They can be thought of as the geometric analog to phonemes, and have been
utilized to create a geometric basis for head models used in commercial 3D
graphics products to parameterize and animate speech~\cite{Zhou2018visemenet}. 
Despite the aforementioned theoretical advantages, visemes have certain practical 
difficulties, often requiring commercial tools such as JALI~\cite{JALI2016}.
Our framework can also handle different frame-level audio feature representations 
such as wav2vec~2.0~\cite{wav2vec2}.

Commonly used intermediate representations, such as keypoints~\cite{Wang18b} can be
inaccurate, especially for difficult-to-detect regions such as lips,
We propose to forego the use of intermediate geometric representation 
and instead render the image directly given visemes.

{\bf Facial features.} To render the rest of the head, landmarks and 
contours are strong guiding signals for the network to produce semantically correct
images. We use these features of the non-mouth areas of the face as additional inputs
for the network, enabling control of non-speech-related behavior such as head motion.

\subsection{Multi-Modal Neural Rendering}\label{sec:viseme2pix}
We start by processing the audio to obtain $k$ features per frame 
$\vc{v} \in \R^{k \times 1}$. We take a window of 6 consecutive frames centered 
around the corresponding image frame. This results in a vector of size 
$\vc{v} \in \R^{k \times 6}$, shown in~\fg{Inference}a. For visemes, $k=34$.

The second modality is the upper face keypoints and the contours drawn on a
canvas of size $512 \times 512$ px, in~\fg{Inference}b.
Rather than passing the keypoints as vectors, we draw them as a 2D
image which helps the model to learn spatial correspondences, resulting in better
generalization to translations and motion in the keypoints. 

{\bf Model architecture details.}
Our model consists of two encoders, one for each modality. The audio encoder 
consists of a 1D convolutional neural network followed by a 2-layer
perceptron network. The resulting latent vector is reshaped to a 2D representation
and concatenated with the latent feature map of the keypoint encoder which consist
of 4-layer strided convolution blocks with a residual block~\cite{He16_residual} in 
between consecutive strided convolutions. Taking the inspiration from 
Pix2PixHD~\cite{Wang18b}, we adopted a two image decoder scheme. One decoder is
trained to generate a full face (\fg{Inference}d), and the second decoder
(\fg{Inference}c) is trained to generate a fixed crop region of the mouth at
a larger resolution. Both decoders produce different images from the same latent space.
The model is trained end to end in a GAN setup, using two patch 
discriminators~\cite{Isola17}: one for the head and the other for the mouth.
We use the VGG~\cite{Simonyan14} feature matching loss and smooth L1 loss alongside
the GAN loss to train the generators. To obtain sharper mouth images, especially for
the teeth, we resize and place the mouth region feature map into the head feature map at
the appropriate crop position in the penultimate layer of the head decoder, as shown
by the red arrow in~\fg{Inference}. We do this in the penultimate layer rather than the
final image to avoid any blending artifacts that could arise.

\subsection{Disentanglement of Audio and Head-Pose}\label{sec:augmentation}
As mentioned in~\sec{viseme2pix}, we use two inputs to the network: 
a $k \times 6$ vector of audio features, and a $512 \times 512$ line drawing of the 
face. Since our training dataset only contains a single subject speaking one 
language, this leads to considerable entanglement between audio features
and the 2D contour drawing, making the model learn unwanted relationships 
between the line drawing and lip motion. This manifests in, for example, lip motion being 
generated from just head motion in the contour drawing, even without 
providing any audio.

To remedy this, we propose a novel data augmentation strategy for disentangling
the audio features from the head pose. One way to enforce the correlation between audio
features and lip shape is to show the network samples of the subject uttering the same
phrases with different head poses. While this can be done during data capture, this significantly
lengthens and complicates the capture process. To remedy this, we propose training an oracle
network that converts 2D contour drawings into photoreal
renders. Then, we propose a method for transferring the mouth shape from one frame to the 
head pose of another. Combining these two methods allows us to augment the capture dataset
with synthetic images containing a greater variety of lip shapes and head poses, forcing the
network to break the correlation between head pose and lip shape.


\begin{figure}
\def\svgwidth{\hsize}
\begingroup%
  \makeatletter%
  \providecommand\color[2][]{%
    \errmessage{(Inkscape) Color is used for the text in Inkscape, but the package 'color.sty' is not loaded}%
    \renewcommand\color[2][]{}%
  }%
  \providecommand\transparent[1]{%
    \errmessage{(Inkscape) Transparency is used (non-zero) for the text in Inkscape, but the package 'transparent.sty' is not loaded}%
    \renewcommand\transparent[1]{}%
  }%
  \providecommand\rotatebox[2]{#2}%
  \newcommand*\fsize{\dimexpr\f@size pt\relax}%
  \newcommand*\lineheight[1]{\fontsize{\fsize}{#1\fsize}\selectfont}%
  \ifx\svgwidth\undefined%
    \setlength{\unitlength}{444bp}%
    \ifx\svgscale\undefined%
      \relax%
    \else%
      \setlength{\unitlength}{\unitlength * \real{\svgscale}}%
    \fi%
  \else%
    \setlength{\unitlength}{\svgwidth}%
  \fi%
  \global\let\svgwidth\undefined%
  \global\let\svgscale\undefined%
  \makeatother%
  \begin{picture}(1,0.32594549)%
    \lineheight{1}%
    \setlength\tabcolsep{0pt}%
    \put(0,0){\includegraphics[width=\unitlength,page=1]{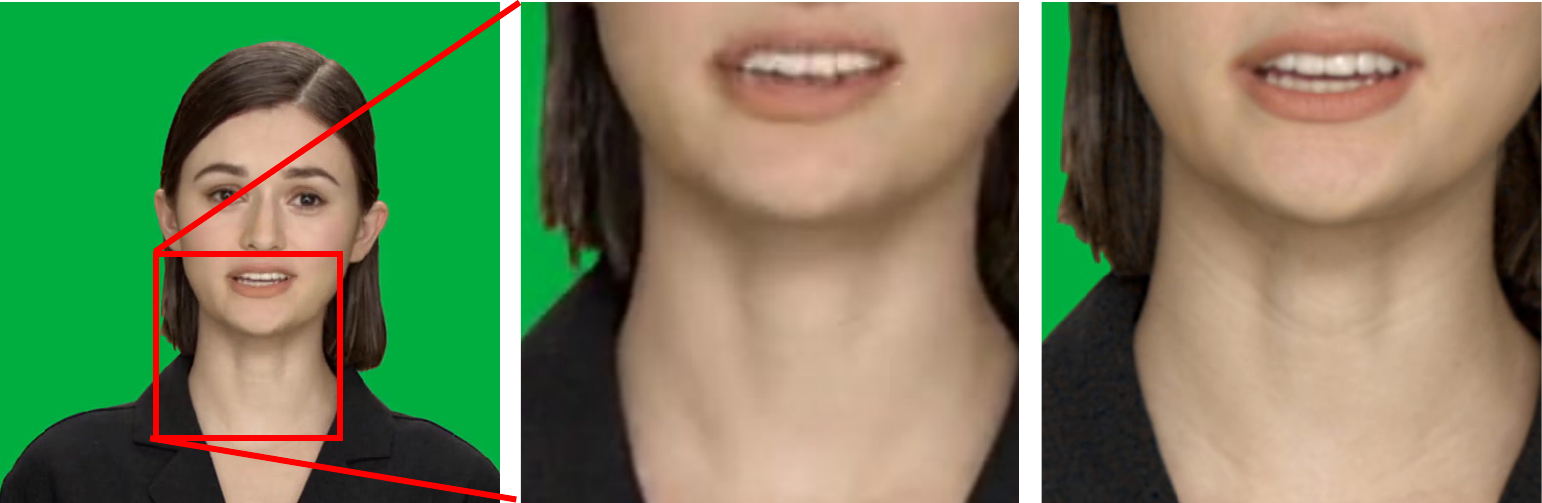}}%
    \put(0.01632524,0.01154688){\color[rgb]{1,1,1}\makebox(0,0)[lt]{\lineheight{1.25}\smash{\begin{tabular}[t]{l}(a) 512 px\end{tabular}}}}%
    \put(0.35380924,0.01154688){\color[rgb]{1,1,1}\makebox(0,0)[lt]{\lineheight{1.25}\smash{\begin{tabular}[t]{l}(b) 192 px\end{tabular}}}}%
    \put(0.69325585,0.01154688){\color[rgb]{1,1,1}\makebox(0,0)[lt]{\lineheight{1.25}\smash{\begin{tabular}[t]{l}(c) 192 px\end{tabular}}}}%
  \end{picture}%
\endgroup%
\caption{
Training a full-head model at 512 px resolution~(a) produces
lower details in the 192 px mouth region~(b) than training
a mouth-only model on 512 px, then resampled to 192px~(c).
}\label{fig:outpaint_motivation}
\end{figure}

{\bf Oracle Network.} The oracle generates photorealistic images of the subject
given the 2D line drawings from in~\fg{Inference}. This can be
accomplished by employing a network based on Pix2Pix~\cite{Isola17,Wang18b}.
While the images produced by this network appear photorealistic, Pix2Pix can
struggle with blurriness, especially in regions where there is significant
motion and occlusions, such as the teeth. This blurriness pollutes our training data,
degrading the visual quality of our method. One observation is that if we train a
higher-resolution network that focuses on the mouth then the teeth will come out clearer,
even if the result is then downsampled as shown in ~\fg{outpaint_motivation}. Here, we trained
two versions of the same network: one on the full-head $512 \times 512$ px
data~\fg{outpaint_motivation}a and the other on high-resolution $512 \times 512$ px mouth crop, then downscaled to $192 \times 192$px ~\fg{outpaint_motivation}c. If we extract
the same $192 \times 192$ region from the full head model (\fg{outpaint_motivation}b), 
we can see that there is a degradation in quality in the teeth region.

To produce high-resolution renderings, we propose to render the 
image in a hierarchical manner starting with the region where the details and 
quality is most important for the given task -- the mouth region in our case.
Our solution can be viewed as a {\em chaining} of neural network models where 
the {\em next} network is conditioned by the result of the {\em previous} network,
shown in~\fg{outpaint_figure}. First, we train a network that generates a high-quality
mouth image(~\fg{outpaint_figure}a). This image is downsampled and placed in a
line drawing of the full head, similar to the process outlined in~\sec{viseme2pix}.
Then, it is used as input to the full-head network(~\fg{outpaint_figure}b), which
learns to render the rest of the head. We refer to this
hierarchical approach as {\em outpaint} since it resembles the task of outpainting.

\begin{figure}[t]
\includegraphics[width=\columnwidth]{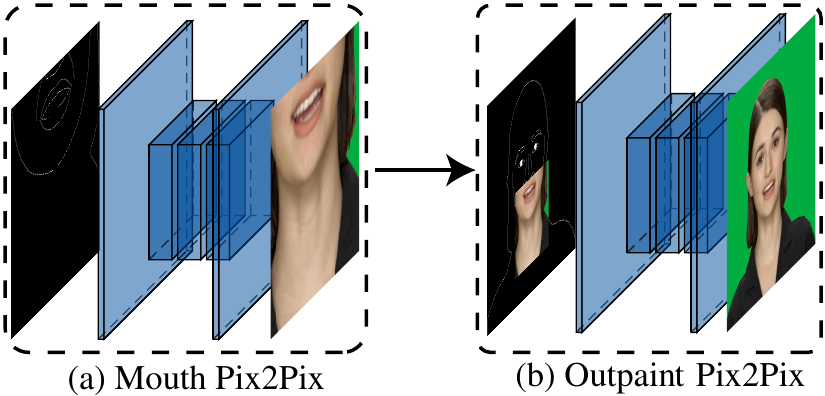}
\caption{
Our oracle network works in a hierarchical fashion, starting with the network to 
generate mouth~(a) in a high-resolution. The result is then fed into the subsequent 
network~(b) to generate the rest of the head. The head image and the high 
resolution mouth image are used for supervision of our multi-modal renderer.  
}
\label{fig:outpaint_figure}
\end{figure}

{\bf Keypoint mashing.} The next missing piece for generating novel head pose
and mouth shape combinations is the
ability to combine the mouth keypoints from one frame with the head keypoints
in another frame. However, the two frames will have different head poses, meaning
we need to remove the global pose information before the mouth keypoints 
can be replaced.

For any given frame, we extracted $n$ 2D facial keypoints 
$\vc{K} \in \R^{n \times 2}$ and a pose matrix $\vc{P} \in \R^{4 \times 4}$
based on a 3D canonical space for the keypoints. The position of
the keypoints can be expressed as:
\begin{align}\label{eq:kp_forward}
\vc{K} = proj(\vc{P} \vc{K^f})
\end{align}
where $\vc{K^f} \in \R^{n \times 4}$ are the un-posed keypoints with depth in 
homogeneous coordinates, and $proj()$ is a projection function (e.g. perspective 
projection) that converts keypoints from 3D space to 2D
screen space. Using~\eq{kp_forward}, we can compute the un-posed keypoints:
\begin{align}\label{eq:kp_inverse}
\vc{K^f} = \vc{P^{-1}}proj^{-1}(\vc{K} )
\end{align}
Where $proj^{-1}$ is the inverse projection function. For an orthographic 
projection, this would just be an inverse of the projection matrix, provided 
the keypoint data is 3D. However, many keypoints detectors only provide 
2D information. In order to fill in the missing depth, we obtain depth
values by solving the forward problem~\eq{kp_forward} for the canonical keypoints
and concatenating the resulting depth value to the extracted 2D keypoints. This 
same process can be used to recover the perspective divide if $proj$ is 
a perspective transformation. While this is only an approximation for the 
true depth values, most of the depth variation in the face comes from the
underlying facial bone structure and geometry captured in the canonical 
keypoints, making this a viable approximation.

\begin{figure}[t]
\includegraphics[width=\columnwidth]{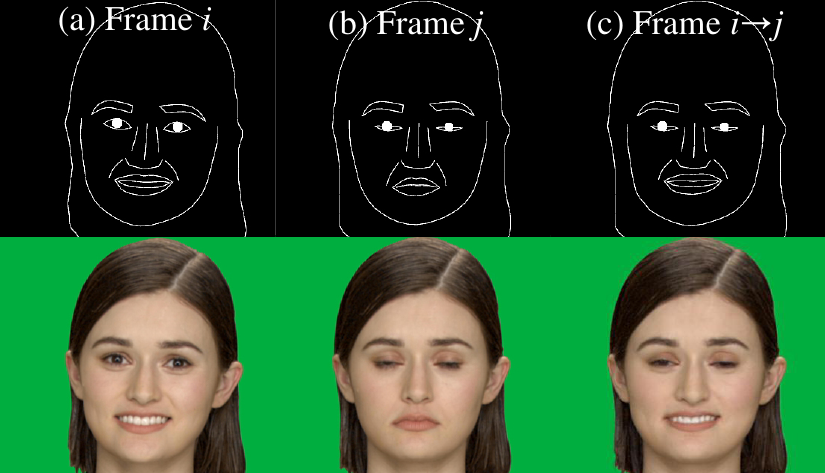}
\caption{
Using keypoint mashing and the proposed oracle network, we are able to generate
a synthetic image~(c) by combining the mouth keypoints of~(a) with the head 
keypoints of~(b).
}
\label{fig:data_augmentation_fig} 
\end{figure}

\begin{figure*}[!h]
\def\svgwidth{\hsize}
\begingroup%
  \makeatletter%
  \providecommand\color[2][]{%
    \errmessage{(Inkscape) Color is used for the text in Inkscape, but the package 'color.sty' is not loaded}%
    \renewcommand\color[2][]{}%
  }%
  \providecommand\transparent[1]{%
    \errmessage{(Inkscape) Transparency is used (non-zero) for the text in Inkscape, but the package 'transparent.sty' is not loaded}%
    \renewcommand\transparent[1]{}%
  }%
  \providecommand\rotatebox[2]{#2}%
  \newcommand*\fsize{\dimexpr\f@size pt\relax}%
  \newcommand*\lineheight[1]{\fontsize{\fsize}{#1\fsize}\selectfont}%
  \ifx\svgwidth\undefined%
    \setlength{\unitlength}{1052.19744873bp}%
    \ifx\svgscale\undefined%
      \relax%
    \else%
      \setlength{\unitlength}{\unitlength * \real{\svgscale}}%
    \fi%
  \else%
    \setlength{\unitlength}{\svgwidth}%
  \fi%
  \global\let\svgwidth\undefined%
  \global\let\svgscale\undefined%
  \makeatother%
  \begin{picture}(1,0.57469018)%
    \lineheight{1}%
    \setlength\tabcolsep{0pt}%
    \put(0,0){\includegraphics[width=\unitlength,page=1]{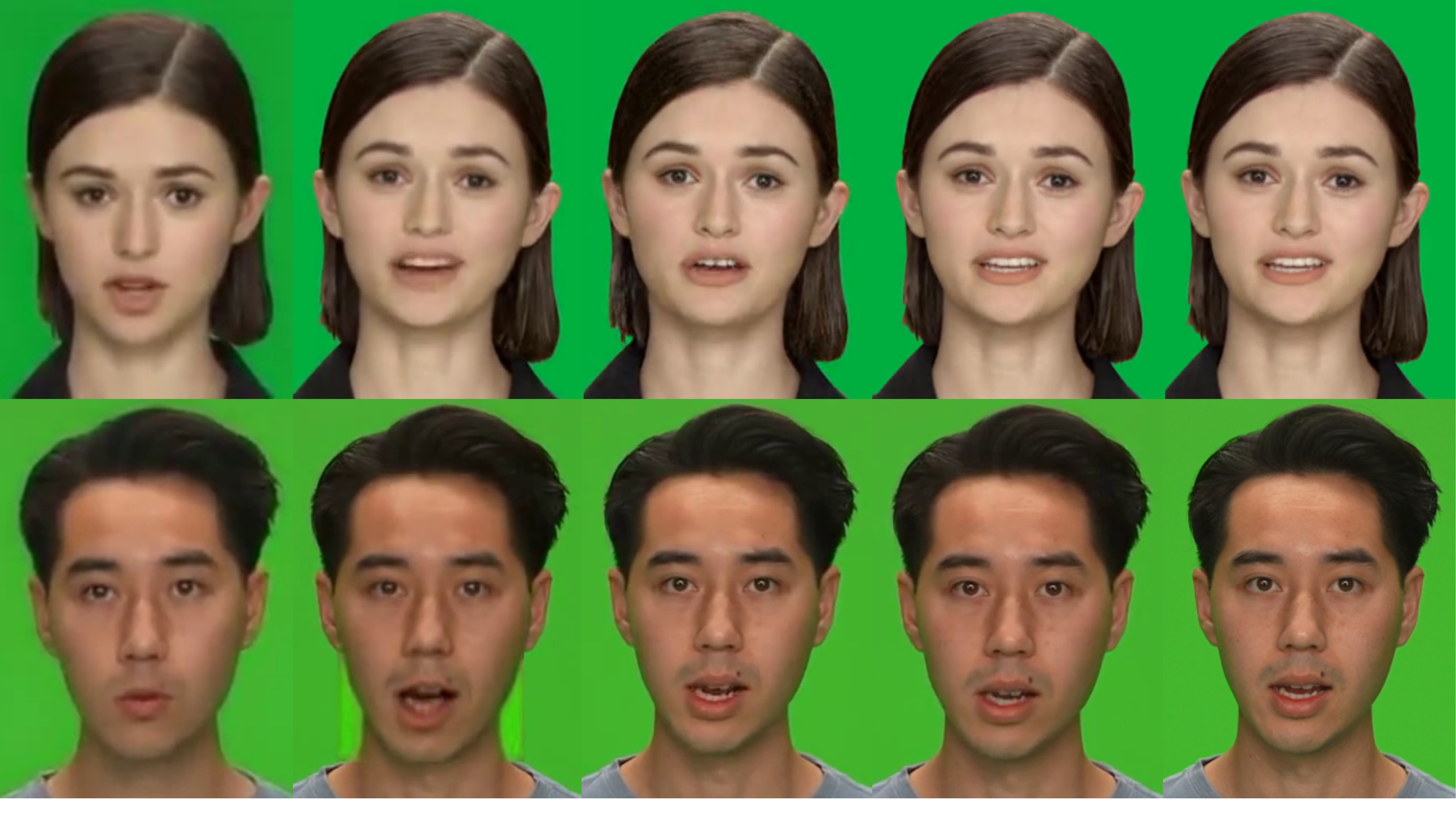}}%
    \put(0.81501653,0.00382383){\color[rgb]{0,0,0}\makebox(0,0)[lt]{\lineheight{1.25}\smash{\begin{tabular}[t]{l}(e) Ground Truth\end{tabular}}}}%
    \put(0.41772177,0.00382383){\color[rgb]{0,0,0}\makebox(0,0)[lt]{\lineheight{1.25}\smash{\begin{tabular}[t]{l}(c) NVP+Pix2Pix\end{tabular}}}}%
    \put(0.24213343,0.00382383){\color[rgb]{0,0,0}\makebox(0,0)[lt]{\lineheight{1.25}\smash{\begin{tabular}[t]{l}(b) Wav2lip\end{tabular}}}}%
    \put(0.0272532,0.00382383){\color[rgb]{0,0,0}\makebox(0,0)[lt]{\lineheight{1.25}\smash{\begin{tabular}[t]{l}(a) MakeItTalk\end{tabular}}}}%
    \put(0.65535016,0.00382383){\color[rgb]{0,0,0}\makebox(0,0)[lt]{\lineheight{1.25}\smash{\begin{tabular}[t]{l}(d) Ours\end{tabular}}}}%
  \end{picture}%
\endgroup%
\caption{
Comparison with state-of-the-art techniques.
MakeItTalk~\cite{Yang2020_MakeItTalk}~(a),
Wav2Lip~\cite{Prajwal2020lip}~(b),
NVP~\cite{Thies2020neural}+Pix2pix~\cite{Wang18}~(c), 
our result~(d), and
ground truth image~(e). 
The avatar says 'ee'. Our results have significantly better lip shapes as 
compared to (a, c), and better texture quality as compared to (a, b).
}\label{fig:results_comparison}
\end{figure*}

{\bf Generating synthetic data.}
With keypoint mashing and a generative oracle in place, we now define a method for
data augmentation. We select two arbitrary frames $i$ and $j$, with corresponding keypoints and poses $\vc{K}_i$,$\vc{P}_i$ (\fg{data_augmentation_fig}a, top) and
$\vc{K}_j$,$\vc{P}_j$ (\fg{data_augmentation_fig}b, top), we compute the un-posed keypoints $\vc{K}^f_i$. Then, we
apply $\vc{P}_j$ resulting in a new keypoints set $\vc{K}_{i\rightarrow j}$: the keypoints
from frame $i$ in the head pose of frame $j$. Since the two keypoint sets $\vc{K}_{i\rightarrow j}$
and $\vc{K}_j$ are now in the same space, we can trivially replace the mouth 
keypoints: $\vc{K}_j^{mouth} = \vc{K}_{i\rightarrow j}^{mouth}$ resulting in the mouth 
position from $i$ transferred over to frame $j$, $\vc{K}_{j}'$. Now we have audio features for
 frame $i$, but a new keypoint set $\vc{K}_{j}'$ (\fg{data_augmentation_fig}c, top).

Using our oracle network, we can now generate synthetic images (\fg{data_augmentation_fig}c, bottom)
from $\vc{K}_{j}'$ to be used as the ``ground truth'' for this new head pose and mouth shape combination.
This makes the network to see different head poses for every audio feature frame, preventing it
from learning an erroneous correlation between the head motion and lip shape.

\section{Results}
\label{sec:results}

First we present comparisons with the current state-of-the-art 
techniques. Then, we elaborate on data collection and 
implementation details, discuss hyper-parameters, followed by further analysis and discussion.
Our method is single identity, and each subject shown has their own model trained on their own dataset.
For more results, please, see our supplementary pdf and video.

\subsection{Comparisons}
{\bf Qualitative comparisons.}
We compare our method with two recent multi-identity approaches,~\fg{results_comparison}a-b:
MakeItTalk~\cite{Yang2020_MakeItTalk} and Wav2Lip~\cite{Prajwal2020lip}. These methods are trained 
on a large corpus of data containing many identities, and extract the identity of the subject at inference time.
While it is indeed practical that MakeItTalk only requires a single image to synthesize a new identity,
the resulting visual and lip sync quality is markedly worse than our method.
Wav2Lip, which infers identity from a source video, arguably produces better lip sync than our method, but at the expense
of visual quality: Wav2Lip only supports $96 \times 96$, of which only
$48 \times 96$ are used for the mouth, and attempting higher resolutions fails. Notice that both of
these above methods produce black blobs in the mouth region without any teeth textures.
We then compared our method to TalkingFace~\cite{Song_2022_CVPR}, which is also
trained on a single identity using data which is similar to our approach making the comparison straightforward. 
Our method produces superior
lip motion: \fg{comparison_cvpr2022}
shows how the two methods compare when producing the word ``we''. Our method is able
to correctly capture the lip transition from the more rounded ``w'' sound to the
``e'' sound. Another single-identity method for generating talking heads is outlined
in Neural Voice Puppetry~\cite{Thies2020neural}. We implemented a variation using Pix2Pix~\cite{Wang18}
and trained it on our data; our method produces superior lip motion~\fg{results_comparison}c-d.
Overall, our method produces higher quality videos with sufficiently accurate lip motion than
the competing methods. Please refer to the supplementary video for more detailed comparisons.

\begin{figure}
\includegraphics[width=\columnwidth]{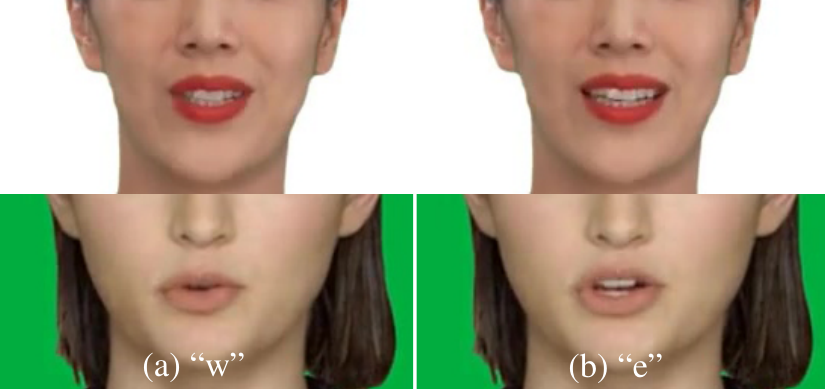}
\caption{
TalkingFace~\cite{Song_2022_CVPR} (top) and our model (bottom) were both asked
to produce the word ``we''. Our model captures the transition from ``w''~(a) 
to ``e''~(b) more accurately. 
}
\label{fig:comparison_cvpr2022}
\end{figure}

{\bf Quantitative comparisons.}
In \tab{Results_table}, we provide quantitative comparisons between our
method and the above competing methods. To evaluate image quality, we compute
the PSNR and SSIM~\cite{SSIM} scores for 500 inferred images against their ground
truth counterparts, and show that we produce the best
results. We also provide two metrics to evaluate the quality of the lip sync:
the score provided by SyncNet~\cite{Chung2016OutOT} (\textbf{Sync}) and Lip Marker Distance (D$_{lip}$) in \tab{Results_table}.
In terms of lip-sync, our results are superior to all of the other comparisons 
except Wav2Lip. This is because Wav2Lip was trained to maximize \textbf{Sync},
which also translates into better D$_{lip}$. Our method comes close to Wav2Lip
in terms of lip sync quality, but with a higher visual quality. Since we do not
have the ground truth videos for Talking Face, we are unable to compute SSIM, PSNR
and D$_{lip}$ metrics, which rely on ground truth frames.

{\bf Performance.}
All results shown in the~\tab{Results_table} were run on a single 
RTX 3080 Ti with a Intel(R) Core(TM) i9-9900K CPU @ 3.60GHz. Our 
method is close to twice as fast as the second fastest method, TalkingFace.
This is because by using a vector of visemes as input, we are able to
greatly simplify our network. Additionally, since our architecture does not
use any custom layers, our models can be easily converted to TensorRT. This
further increases our inference speeds up to 200 frames per second, making it
very suitable for production-level interactive applications.

\begin{table}
	\begin{center}		
		\begin{tabular}{| c |c | c| c| c| c| c|}
			\toprule
			\textbf{Method} & \shortstack{ \textbf{FPS}  \\ $\uparrow$ } & \shortstack{ \textbf{SSIM} \\ $\uparrow$ } & \shortstack {\textbf{PSNR} \\ $\uparrow$ } & \shortstack {\textbf{Sync} \\ $\uparrow$} & \shortstack{\textbf{D}$_{lip}$ \\ $\downarrow$ } \\
			\toprule
			Wav2Lip  &  49 &  0.88 &  33.16 &  \bf 9.51 & \textbf{8.65}\\
			\hline
			TalkingFace & 63 & na & na & 6.45 &  na \\
			\hline 
			NVP+p2p & 47 & 0.82 & 27.19 & 4.83 &  18.34\\
			\hline
			{\bf Ours} & \bf 110 & \textbf{0.94} & \textbf{35.71} & 7.41 &  12.81\\ 
			\bottomrule
		\end{tabular}
		\caption{\label{tab:Results_table} Quantitative comparisons with 
			Wav2Lip\cite{ Prajwal2020lip}, TalkingFace\cite{Song_2022_CVPR}
			and NVP\cite{Thies2020neural}+Pix2Pix\cite{Wang18}. 
			Sync confidence score \textbf{Sync} and Lip 
			Distance metric \textbf{D$_{lip}$} are used to determine accuracy of lip synchronization. 
			Structure similarity score \textbf{SSIM} and peak signal to noise ration \textbf{PSNR} 
			are used to judge the image quality. Our methods outperform other methods by a 
			large margin on image quality and trails close behind Wav2Lip in lip-sync accuracy 
			while achieving the highest inference speed \textbf{FPS} by a significant margin.}
	\end{center}
\end{table}

\subsection{Discussion and Analysis}

{\bf Data Collection and Capture Setup.}
We recorded subjects using cameras capable of capturing 6K footage at 30 
frames per second. From the full-sized video, we crop out the head region
and a high-resolution mouth image that is used to supervise the training. We
then run facial landmark detection, pose estimation, and contour extraction
on the head crops, which provides us with the information we need to produce 
line-drawings such as the ones in~\fg{Inference}. For audio data, we extract 
visemes by providing the audio waveforms and a transcript to JALI~\cite{JALI2016}.

{\bf Implementation Details.}
We detect facial keypoints and head pose using a commercial product AlgoFace 
(www.algoface.ai). To obtain contours, we first generate a head segmentation 
using a BiSeNetV2~\cite{Yu2021_BiSeNet} and then we draw a line on the 
background-foreground boundary. Finally, we use JALI~\cite{JALI2016} to extract visemes.
The viseme window size is a hyper-parameter; we empirically determined that using a window
of size 6 is optimal. Larger window sizes lead to muffled mouth motion and very small window
sizes produces unstable results. For the generator and both of the discriminators, we use Adam
optimizer with a learning rate of $0.0002$ and learning rate scheduling. We train the system
on a 20 minute video of the subject speaking and evaluate on out of domain TTS audio paired
 with arbitrary sequences of keypoints. The oracle and multi-modal renderer each take
roughly 6-8 hours on two 3080 GPUs with a batch size of 1.

\begin{figure}[t]
\def\svgwidth{\hsize}
\begingroup%
  \makeatletter%
  \providecommand\color[2][]{%
    \errmessage{(Inkscape) Color is used for the text in Inkscape, but the package 'color.sty' is not loaded}%
    \renewcommand\color[2][]{}%
  }%
  \providecommand\transparent[1]{%
    \errmessage{(Inkscape) Transparency is used (non-zero) for the text in Inkscape, but the package 'transparent.sty' is not loaded}%
    \renewcommand\transparent[1]{}%
  }%
  \providecommand\rotatebox[2]{#2}%
  \newcommand*\fsize{\dimexpr\f@size pt\relax}%
  \newcommand*\lineheight[1]{\fontsize{\fsize}{#1\fsize}\selectfont}%
  \ifx\svgwidth\undefined%
    \setlength{\unitlength}{406.50013733bp}%
    \ifx\svgscale\undefined%
      \relax%
    \else%
      \setlength{\unitlength}{\unitlength * \real{\svgscale}}%
    \fi%
  \else%
    \setlength{\unitlength}{\svgwidth}%
  \fi%
  \global\let\svgwidth\undefined%
  \global\let\svgscale\undefined%
  \makeatother%
  \begin{picture}(1,0.32841326)%
    \lineheight{1}%
    \setlength\tabcolsep{0pt}%
    \put(0,0){\includegraphics[width=\unitlength,page=1]{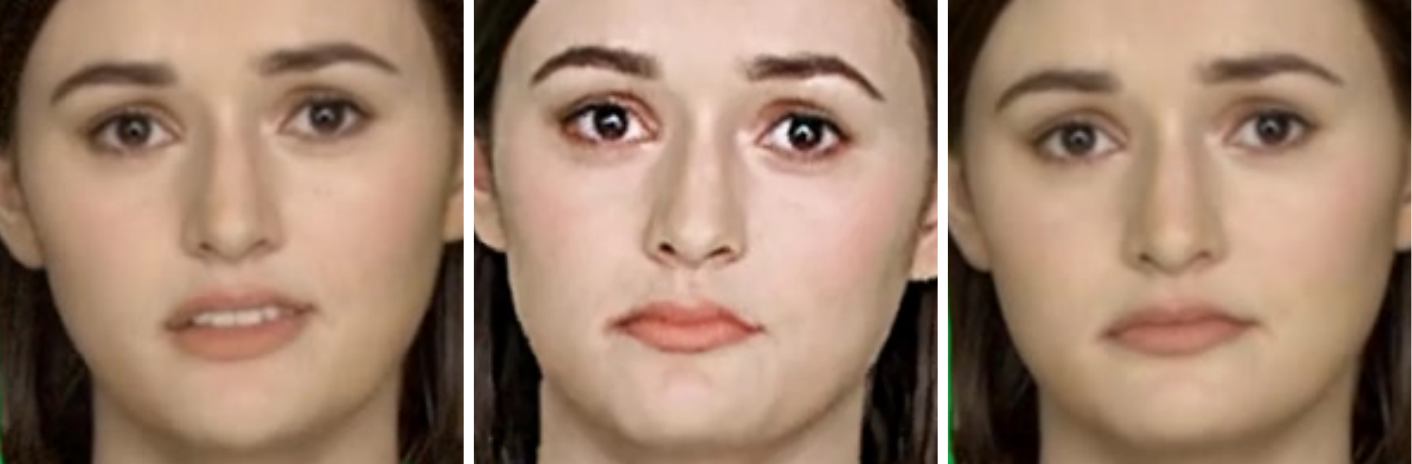}}%
    \put(0.6787032,0.02018462){\color[rgb]{1,1,1}\makebox(0,0)[lt]{\lineheight{1.25}\smash{\begin{tabular}[t]{l}(c) Our aug.\end{tabular}}}}%
    \put(0.0098534,0.02018662){\color[rgb]{1,1,1}\makebox(0,0)[lt]{\lineheight{1.25}\smash{\begin{tabular}[t]{l}(a) Without aug.\end{tabular}}}}%
    \put(0.34195009,0.02018662){\color[rgb]{1,1,1}\makebox(0,0)[lt]{\lineheight{1.25}\smash{\begin{tabular}[t]{l}(b) Classical aug.\end{tabular}}}}%
  \end{picture}%
\endgroup%
\caption{
The proposed augmentation is essential for the model to correctly respect 
different inputs. Without the augmentation~(a), the virtual human erroneously 
opens her mouth even when the audio is silence; 
while (b) classical data augmentation methods might slightly mitigate this issue 
the model still produces wrong lip shapes;
this is fixed by our augmentation technique~(c).
}\label{fig:disentanglement_ablation}
\label{fig:disentanglement}
\end{figure}

{\bf Disentanglement.}
Our network takes two types of input: speech data represented as visemes and
head motion data represented as keypoint drawings. These two modalities are inherently
connected: humans move their heads rhythmically while talking in a way that
matches their speech, e.g., nodding up and down subtly when saying ``yes''.
Without any special handling, our network learns a strong
correlation between the head motion and word being spoken. This results in the
model producing mouth openings just from head motion, depicted in~\fg{disentanglement}a.
Generating synthetic data using classical data augmentation techniques such 
as rotation, scale, translation, cropping, or adding noise would not be effective 
in our case. While it is true that it might slightly help with the overfitting 
as it introduces randomness to the data and it might have effect similar to 
regularization, this alone does not break the correlation between head 
motion and audio, see~\fg{disentanglement}b and our supplementary video, the lips still 
move and the mouth is not in the neutral resting pose.
Our data augmentation successfully breaks the correlation
by introducing a variety of head pose and viseme combinations. \fg{disentanglement}c
shows that our model properly keeps its mouth closed during silence, regardless of head motion.
We empirically observed that at least 80\% of the training data have to be synthetic 
data for the disentanglement to have desired effect.

\begin{figure}[!h]
\includegraphics[width=\columnwidth]{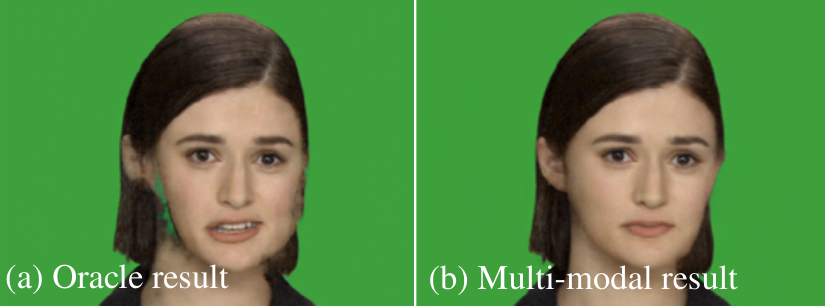}
\caption{
Even though the oracle~(a) can sometimes produce artifacts in our synthetic 
training data, our multimodal renderer~(b) is robust to them.}
\label{fig:oracle_artifact}
\end{figure}

Since the keypoint mashing \sec{augmentation} can produce unnatural combinations that
are substantially different from the oracle's training data, the synthetic data
produced can contain noticeable artifacts (\fg{oracle_artifact}a). However, such
artifacts only occur when the two frames have vastly different head poses; this is
rare and results in only a small percentage of the training data being polluted. Our
multi-modal network is robust to such artifacts, shown in \fg{oracle_artifact}b.

{\bf Different backbones and performance optimization.}
MobileNetV2~\cite{MobileNetV2} is usually used as compressed architectures primarily for 
image recognition tasks. We tried to directly apply the MobileNetV2 backbone for our image 
synthesis tasks and observed that both {\em large} and {\em small} variants of the MobileNetV2 were slower than our 
current approach. Using hard swish activation as proposed in MobileNetV2 led to rendering 
artifacts across multiple runs, and we chose to use the original leaky relu activation functions.
Finally, replacing all convolutions in our network with depth wise separable convolutions 
did improve the performance, but also introduced some quality degradation in the teeth and 
produced grid like artifacts near high contrast texture regions like the hair.

\section{Limitations and Future Work}
\label{sec:limits}
While our work significantly advances the state-of-the-art in generating 
realistic virtual human avatars, there are still certain limitations and a 
fair amount of potential future work that needs to be done before the 
virtual humans are truly indistinguishable from real people.

One of the major limitations of our framework is that it does not perform well on
large motions, head rotations, and extreme head poses.  
To alleviate this in a future work, we envision involving 3D geometry and 
other aspects from the 3D rendering domain, in particular, using a mesh 
as an intermediate representation instead of 2D images, which would allow
for better occlusion and collision handling and for using 3D neural rendering
techniques, such as deferred neural rendering~\cite{Thies2019DNR} that have been proven
very successful.

Furthermore, we also observed certain texture-sticking artifacts between frames
when the motion is large, this is due to the fully convolutional nature of our 
network; this is a known problem discussed in, for example, 
StyleGAN3~\cite{Karras2021}, and might be mitigated by adopting vision 
transfomers~\cite{Lee2022_vitgan}.
Also, as a part of further work, we encourage exploration into how 
multiple modalities can be used to target the same part of the face, enabling 
us to selectively learn from each of the modalities~\cite{huang2022poegan}.
In the context of this work, this could mean, for example, having visemes 
as one modality and a signal to controls smiling as another modality;
both modalities used to synthesize lips.

\section{Conclusion}
\label{sec:conclude}
While the problem of creating virtual humans that perfectly mimic the 
appearance and behavior of real people is still far from being solved,
our work considerably pushes the state-of-the-art boundaries.
We presented a robust and efficient framework for generating photorealistic 
talking face animations from audio in real-time. Thanks to the proposed alternative 
data representation, a training system to prevent modality entanglement, and 
supervision from high resolution around the mouth area, we are able to produce 
superior face rendering quality with better lip synchronization compared to
recent approaches, all while maintaining real-time inference.

\section*{Acknowledgments}
We would like to thank 
Sajid Sadi, Ankur Gupta, Anthony Liot, Anil Unnikrishnan, Janvi Palan, 
and rest of the team at \href{https://neonlife.ai}{NEON} for their extensive
engineering, hardware, and design support.


\clearpage
\appendix


\section{Additional Results}
In~\fg{results_big} we show additional results on a diverse set of identities, 
three females and two males, all of a different ethnicity and skin color. 
As demonstrated by the second row, our method can handle {\em floppy} hair.
Also, it is able to faithfully preserve the facial hair, see the last row.

\section{Mouth Decoder Ablation Study.}
Human faces have a great range of different frequencies of motion in different 
face regions. In order to accurately capture these various frequencies, we 
task the head and mouth decoders of our network to learn different regions 
of the face from the same latent space, offering us higher image quality 
and correctness for generating the head and mouth images respectively. We 
conduct an ablation study to show that without this 2-decoder arrangement 
and feature pasting, the quality of images generated by the head generator 
is inferior~\fg{decoder_ablation}a compared to~\fg{decoder_ablation}b.

\begin{figure}[h]
	\def\svgwidth{\hsize}
\begingroup%
  \makeatletter%
  \providecommand\color[2][]{%
    \errmessage{(Inkscape) Color is used for the text in Inkscape, but the package 'color.sty' is not loaded}%
    \renewcommand\color[2][]{}%
  }%
  \providecommand\transparent[1]{%
    \errmessage{(Inkscape) Transparency is used (non-zero) for the text in Inkscape, but the package 'transparent.sty' is not loaded}%
    \renewcommand\transparent[1]{}%
  }%
  \providecommand\rotatebox[2]{#2}%
  \newcommand*\fsize{\dimexpr\f@size pt\relax}%
  \newcommand*\lineheight[1]{\fontsize{\fsize}{#1\fsize}\selectfont}%
  \ifx\svgwidth\undefined%
    \setlength{\unitlength}{257.55631256bp}%
    \ifx\svgscale\undefined%
      \relax%
    \else%
      \setlength{\unitlength}{\unitlength * \real{\svgscale}}%
    \fi%
  \else%
    \setlength{\unitlength}{\svgwidth}%
  \fi%
  \global\let\svgwidth\undefined%
  \global\let\svgscale\undefined%
  \makeatother%
  \begin{picture}(1,0.50582406)%
    \lineheight{1}%
    \setlength\tabcolsep{0pt}%
    \put(0,0){\includegraphics[width=\unitlength,page=1]{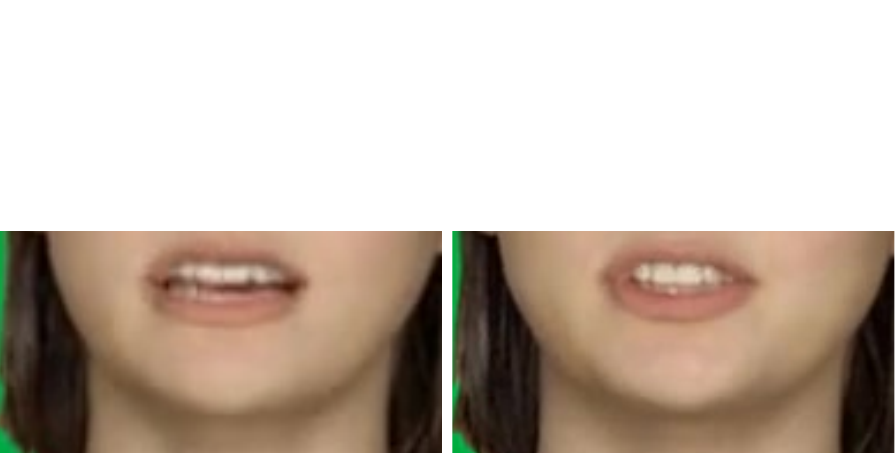}}%
    \put(0.01266753,0.02218255){\color[rgb]{1,1,1}\makebox(0,0)[lt]{\lineheight{1.25}\smash{\begin{tabular}[t]{l}(a) 1 decoder\end{tabular}}}}%
    \put(0.51935284,0.02218255){\color[rgb]{1,1,1}\makebox(0,0)[lt]{\lineheight{1.25}\smash{\begin{tabular}[t]{l}(b) 2 decoders\end{tabular}}}}%
    \put(0,0){\includegraphics[width=\unitlength,page=2]{decoder_ablation_x.pdf}}%
  \end{picture}%
\endgroup%
\caption{
		Image quality of the mouth region produced by the two decoder model~(b) 
		is superior compared to a single decoder model~(a). Teeth in~(a) do not have
		correct shape and appear blurry.
	}\label{fig:decoder_ablation}
\end{figure}
\begin{figure}[h]
	\def\svgwidth{\hsize}
\begingroup%
  \makeatletter%
  \providecommand\color[2][]{%
    \errmessage{(Inkscape) Color is used for the text in Inkscape, but the package 'color.sty' is not loaded}%
    \renewcommand\color[2][]{}%
  }%
  \providecommand\transparent[1]{%
    \errmessage{(Inkscape) Transparency is used (non-zero) for the text in Inkscape, but the package 'transparent.sty' is not loaded}%
    \renewcommand\transparent[1]{}%
  }%
  \providecommand\rotatebox[2]{#2}%
  \newcommand*\fsize{\dimexpr\f@size pt\relax}%
  \newcommand*\lineheight[1]{\fontsize{\fsize}{#1\fsize}\selectfont}%
  \ifx\svgwidth\undefined%
    \setlength{\unitlength}{256.65529633bp}%
    \ifx\svgscale\undefined%
      \relax%
    \else%
      \setlength{\unitlength}{\unitlength * \real{\svgscale}}%
    \fi%
  \else%
    \setlength{\unitlength}{\svgwidth}%
  \fi%
  \global\let\svgwidth\undefined%
  \global\let\svgscale\undefined%
  \makeatother%
  \begin{picture}(1,0.43146747)%
    \lineheight{1}%
    \setlength\tabcolsep{0pt}%
    \put(0,0){\includegraphics[width=\unitlength,page=1]{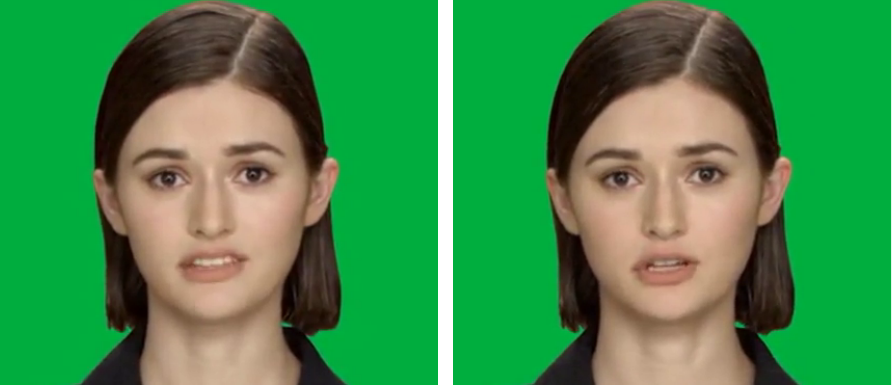}}%
    \put(0.01589177,0.02173334){\color[rgb]{1,1,1}\makebox(0,0)[lt]{\lineheight{1.25}\smash{\begin{tabular}[t]{l}(a) visemes\end{tabular}}}}%
    \put(0.52435586,0.02173334){\color[rgb]{1,1,1}\makebox(0,0)[lt]{\lineheight{1.25}\smash{\begin{tabular}[t]{l}(b) wav2vec\end{tabular}}}}%
  \end{picture}%
\endgroup%
\caption{
		Our system is not limited to any particular audio representation and can work, for example, 
		with visemes~(a) and wav2vec~(b). While the two images look slightly different even though they 
		are supposed to pronounce the same sound, both are a correct representation of the sound 'ee'. 
	}\label{fig:ablation_visemes_vs_wav2vec}
\end{figure}

\section{Alternative Audio Representations}
Since visemes can be difficult and impractical to obtain, we tested our method 
with alternative audio representations. 
Our system trained using visemes~\fg{ablation_visemes_vs_wav2vec}a 
produces comparable results to one trained using
wav2vec~\fg{ablation_visemes_vs_wav2vec}b.
This is because both visemes and wav2vec audio representations exhibit 
similar properties: both are frame-level features with an adequate degree of 
voice and language independence.
Please refer to the supplementary material for further discussion.

\section{Representation of Input Data}
\fg{input_representation} shows an illustration of the input features our
system uses. The contour drawing is represented as a one channel image,
the visemes are represented as a 1D vector.

\begin{figure}[!h]
	\def\svgwidth{\hsize}
\begingroup%
  \makeatletter%
  \providecommand\color[2][]{%
    \errmessage{(Inkscape) Color is used for the text in Inkscape, but the package 'color.sty' is not loaded}%
    \renewcommand\color[2][]{}%
  }%
  \providecommand\transparent[1]{%
    \errmessage{(Inkscape) Transparency is used (non-zero) for the text in Inkscape, but the package 'transparent.sty' is not loaded}%
    \renewcommand\transparent[1]{}%
  }%
  \providecommand\rotatebox[2]{#2}%
  \newcommand*\fsize{\dimexpr\f@size pt\relax}%
  \newcommand*\lineheight[1]{\fontsize{\fsize}{#1\fsize}\selectfont}%
  \ifx\svgwidth\undefined%
    \setlength{\unitlength}{315bp}%
    \ifx\svgscale\undefined%
      \relax%
    \else%
      \setlength{\unitlength}{\unitlength * \real{\svgscale}}%
    \fi%
  \else%
    \setlength{\unitlength}{\svgwidth}%
  \fi%
  \global\let\svgwidth\undefined%
  \global\let\svgscale\undefined%
  \makeatother%
  \begin{picture}(1,0.53082297)%
    \lineheight{1}%
    \setlength\tabcolsep{0pt}%
    \put(0.0079049,0.01390458){\color[rgb]{0,0,0}\makebox(0,0)[lt]{\lineheight{1.25}\smash{\begin{tabular}[t]{l}(a) conotur drawing\end{tabular}}}}%
    \put(0.53659433,0.00894097){\color[rgb]{0,0,0}\makebox(0,0)[lt]{\lineheight{1.25}\smash{\begin{tabular}[t]{l}(b) visemes\end{tabular}}}}%
    \put(0,0){\includegraphics[width=\unitlength,page=1]{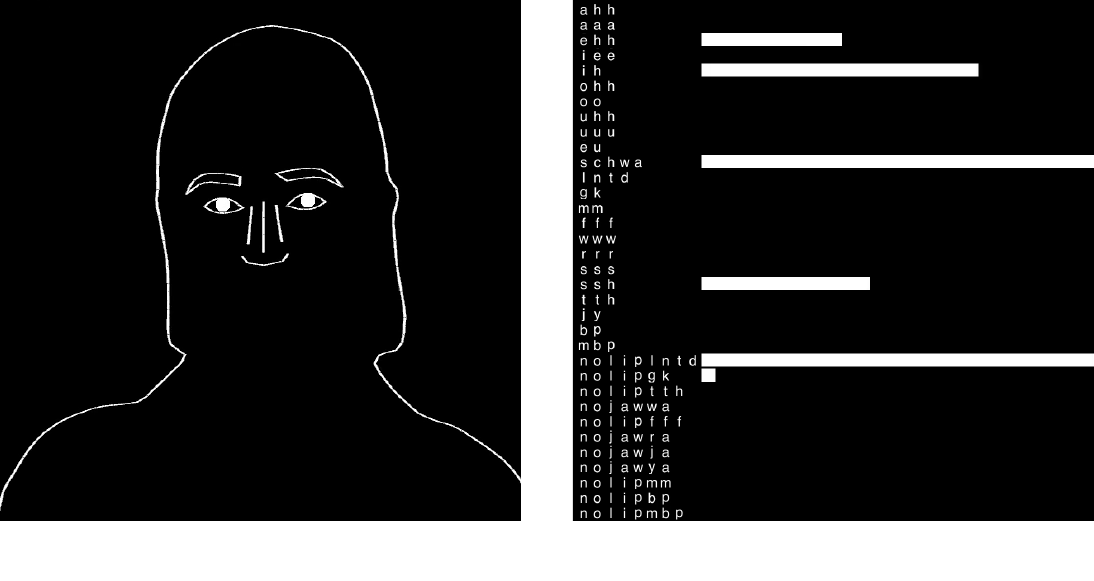}}%
  \end{picture}%
\endgroup%
\caption{
		Visualization of the input data. 
		(a)~contour drawing, 
		(b)~visualization of visemes.
	}\label{fig:input_representation}
\end{figure}

\section{Voice and Language Independence}
Since visemes are geometric shapes, they are largely independent of the voice 
characteristics. Because of this, our model, trained on a single voice, is 
able to produce good lip motion on a variety of voices, including natural 
voices from other people and synthetic voices from, e.g., text-to-speech. 
For the same reasons, visemes also provide some level of language independence.
However, different languages can have very different phonetic distributions 
in common speech. When considering two languages that have similar 
distributions, such as English and Spanish, our English-trained model 
performs well on Spanish audio. However, when we use Korean audio with the 
same model, there is a degradation in the lip sync quality. These results 
are shown in the supplementary video.

\section{Network Architecture Details}
Our network consist of several building blocks. It comprises of two encoders, one for 
contour drawings~\tab{Arch}a and one for audio~\tab{Arch}b. Fusion block~\tab{Arch}c to
combine the two modalities. We also utilize two decoders, one for synthesizing the 
mouth region~\tab{Arch}d and the other to supervise synthesis of the entire head 
region~\tab{Arch}e.

\begin{table}
	\begin{center}		
		\scriptsize
		
		\begin{tabular}{|llllllll|}
			\hline
			\multicolumn{1}{|l|}{\textbf{Layer}} & \multicolumn{1}{l|}{\textbf{W}} & \multicolumn{1}{l|}{\textbf{S}} & \multicolumn{1}{l|}{\textbf{IP}} & \multicolumn{1}{l|}{\textbf{OP}} & \multicolumn{1}{l|}{\textbf{$C_{in}$}} & \multicolumn{1}{l|}{\textbf{$C_{out}$}} & \textbf{Output}    \\ \hline
			\multicolumn{8}{|c|}{\textbf{(a) Keypoint Encoder}}                                                                                                                                                                                                                                                                                   \\ \hline
			\multicolumn{1}{|l|}{Conv}      & \multicolumn{1}{l|}{3}          & \multicolumn{1}{l|}{2}          & \multicolumn{1}{l|}{1}           & \multicolumn{1}{l|}{}            & \multicolumn{1}{l|}{3}                  & \multicolumn{1}{l|}{8}                                   & $8 \times 256 \times 256$          \\ \hline
			\multicolumn{1}{|l|}{Conv}      & \multicolumn{1}{l|}{3}          & \multicolumn{1}{l|}{2}          & \multicolumn{1}{l|}{1}           & \multicolumn{1}{l|}{}            & \multicolumn{1}{l|}{8}                  & \multicolumn{1}{l|}{16}                                  & $16 \times 128 \times 128$         \\ \hline
			\multicolumn{1}{|l|}{Conv}      & \multicolumn{1}{l|}{3}          & \multicolumn{1}{l|}{2}          & \multicolumn{1}{l|}{1}           & \multicolumn{1}{l|}{}            & \multicolumn{1}{l|}{16}                 & \multicolumn{1}{l|}{32}                                  & $32 \times 64 \times 64$           \\ \hline
			\multicolumn{1}{|l|}{Conv}      & \multicolumn{1}{l|}{3}          & \multicolumn{1}{l|}{2}          & \multicolumn{1}{l|}{1}           & \multicolumn{1}{l|}{}            & \multicolumn{1}{l|}{32}                 & \multicolumn{1}{l|}{64}                                  & \textbf{$64 \times 32 \times 32$}  \\ \hline
			\multicolumn{8}{|c|}{\textbf{(b) Audio Encoder}}                                                                                                                                                                                                                                                                                      \\ \hline
			\multicolumn{1}{|l|}{Conv1d}         & \multicolumn{1}{l|}{1}          & \multicolumn{1}{l|}{1}          & \multicolumn{1}{l|}{0}           & \multicolumn{1}{l|}{}            & \multicolumn{1}{l|}{6}                  & \multicolumn{1}{l|}{1}                                 & $1 \times 34$               \\ \hline
			\multicolumn{1}{|l|}{MLP}         & \multicolumn{1}{l|}{}           & \multicolumn{1}{l|}{}           & \multicolumn{1}{l|}{}            & \multicolumn{1}{l|}{}            & \multicolumn{1}{l|}{34}                 & \multicolumn{1}{l|}{1024}                              & 1024               \\ \hline
			\multicolumn{1}{|l|}{MLP}         & \multicolumn{1}{l|}{}           & \multicolumn{1}{l|}{}           & \multicolumn{1}{l|}{}            & \multicolumn{1}{l|}{}            & \multicolumn{1}{l|}{1024}               & \multicolumn{1}{l|}{1024}                               & 1024               \\ \hline
			\multicolumn{1}{|l|}{Reshape}        & \multicolumn{1}{l|}{}           & \multicolumn{1}{l|}{}           & \multicolumn{1}{l|}{}            & \multicolumn{1}{l|}{}            & \multicolumn{1}{l|}{}                   & \multicolumn{1}{l|}{}                                   & \textbf{$1 \times 32 \times 32$}   \\ \hline
			\multicolumn{8}{|c|}{\textbf{(c) Fusion}}                                                                                                                                                                                                                                                                                             \\ \hline
			\multicolumn{1}{|l|}{Cat}    & \multicolumn{1}{l|}{}           & \multicolumn{1}{l|}{}           & \multicolumn{1}{l|}{}            & \multicolumn{1}{l|}{}            & \multicolumn{1}{l|}{}                   & \multicolumn{1}{l|}{}                                   & $65 \times 32 \times 32$           \\ \hline
			\multicolumn{1}{|l|}{Conv}      & \multicolumn{1}{l|}{3}          & \multicolumn{1}{l|}{1}          & \multicolumn{1}{l|}{1}           & \multicolumn{1}{l|}{}            & \multicolumn{1}{l|}{65}                 & \multicolumn{1}{l|}{64}                                 & \textbf{$ 64 \times 32 \times 32$}  \\ \hline
			\multicolumn{8}{|c|}{\textbf{(d) Mouth Decoder}}                                                                                                                                                                                                                                                                                      \\ \hline
			\multicolumn{1}{|l|}{TConv}     & \multicolumn{1}{l|}{3}          & \multicolumn{1}{l|}{2}          & \multicolumn{1}{l|}{1}           & \multicolumn{1}{l|}{1}           & \multicolumn{1}{l|}{64}                 & \multicolumn{1}{l|}{32}                                 & $32 \times 64 \times 64$           \\ \hline
			\multicolumn{1}{|l|}{TConv}     & \multicolumn{1}{l|}{3}          & \multicolumn{1}{l|}{2}          & \multicolumn{1}{l|}{1}           & \multicolumn{1}{l|}{1}           & \multicolumn{1}{l|}{32}                 & \multicolumn{1}{l|}{16}                                & $16 \times 128 \times 128$         \\ \hline
			\multicolumn{1}{|l|}{TConv}     & \multicolumn{1}{l|}{3}          & \multicolumn{1}{l|}{2}          & \multicolumn{1}{l|}{1}           & \multicolumn{1}{l|}{1}           & \multicolumn{1}{l|}{16}                 & \multicolumn{1}{l|}{8}                                & $8 \times 256 \times 256$          \\ \hline
			\multicolumn{1}{|l|}{TConv}     & \multicolumn{1}{l|}{3}          & \multicolumn{1}{l|}{2}          & \multicolumn{1}{l|}{1}           & \multicolumn{1}{l|}{1}           & \multicolumn{1}{l|}{8}                  & \multicolumn{1}{l|}{8}                                 & $8 \times 512 \times 512$          \\ \hline
			\multicolumn{1}{|l|}{Conv}      & \multicolumn{1}{l|}{3}          & \multicolumn{1}{l|}{1}          & \multicolumn{1}{l|}{1}           & \multicolumn{1}{l|}{}            & \multicolumn{1}{l|}{8}                  & \multicolumn{1}{l|}{8}                                & $8 \times 512 \times 512$          \\ \hline
			\multicolumn{1}{|l|}{Conv}      & \multicolumn{1}{l|}{7}          & \multicolumn{1}{l|}{1}          & \multicolumn{1}{l|}{3}           & \multicolumn{1}{l|}{}            & \multicolumn{1}{l|}{8}                  & \multicolumn{1}{l|}{3}                              & \textbf{$3 \times 512 \times 512$} \\ \hline
			\multicolumn{8}{|c|}{\textbf{(e) Face Decoder}}                                                                                                                                                                                                                                                                                       \\ \hline
			\multicolumn{1}{|l|}{TConv}     & \multicolumn{1}{l|}{3}          & \multicolumn{1}{l|}{2}          & \multicolumn{1}{l|}{1}           & \multicolumn{1}{l|}{1}           & \multicolumn{1}{l|}{64}                 & \multicolumn{1}{l|}{32}                              & $32 \times 64 \times 64$           \\ \hline
			\multicolumn{1}{|l|}{TConv}     & \multicolumn{1}{l|}{3}          & \multicolumn{1}{l|}{2}          & \multicolumn{1}{l|}{1}           & \multicolumn{1}{l|}{1}           & \multicolumn{1}{l|}{32}                 & \multicolumn{1}{l|}{16}                           & $16 \times 128 \times 128$         \\ \hline
			\multicolumn{1}{|l|}{TConv}     & \multicolumn{1}{l|}{3}          & \multicolumn{1}{l|}{2}          & \multicolumn{1}{l|}{1}           & \multicolumn{1}{l|}{1}           & \multicolumn{1}{l|}{16}                 & \multicolumn{1}{l|}{8}                             & $8 \times 256 \times 256$          \\ \hline
			\multicolumn{1}{|l|}{TConv}     & \multicolumn{1}{l|}{3}          & \multicolumn{1}{l|}{2}          & \multicolumn{1}{l|}{1}           & \multicolumn{1}{l|}{1}           & \multicolumn{1}{l|}{8}                  & \multicolumn{1}{l|}{8}                             & $8 \times 512 \times 512$          \\ \hline
			\multicolumn{1}{|l|}{Conv}      & \multicolumn{1}{l|}{3}          & \multicolumn{1}{l|}{1}          & \multicolumn{1}{l|}{1}           & \multicolumn{1}{l|}{}            & \multicolumn{1}{l|}{8}                  & \multicolumn{1}{l|}{8}                                 & $8 \times 512 \times 512$          \\ \hline
			\multicolumn{1}{|l|}{Conv}      & \multicolumn{1}{l|}{7}          & \multicolumn{1}{l|}{1}          & \multicolumn{1}{l|}{3}           & \multicolumn{1}{l|}{}            & \multicolumn{1}{l|}{8}                  & \multicolumn{1}{l|}{3}                                  & \textbf{$3 \times 512 \times 512$} \\ \hline
		\end{tabular}
		\caption{\label{tab:Arch} 
			Details of the architecture of neural network blocks used in our framework. 
			\textbf{Conv} comprises of a Convolution2D layer, Instance Normalization, and a LeakyReLU(0.2) activation. 
			\textbf{TConv} comprises of a TransposeConvolution2D layer, Instance Normalization, and LeakyReLU(0.2) activation.
			\textbf{W} is a kernel size, \textbf{S} is a stride, \textbf{IP} is an input padding, \textbf{OP} is an output 
			padding, $C_{in}$ is number of input channels, and $C_{out}$ is number of output channels.
		}
	\end{center}
\end{table}

\begin{figure*}
	\def\svgwidth{\hsize}\import{figs/results_big/}{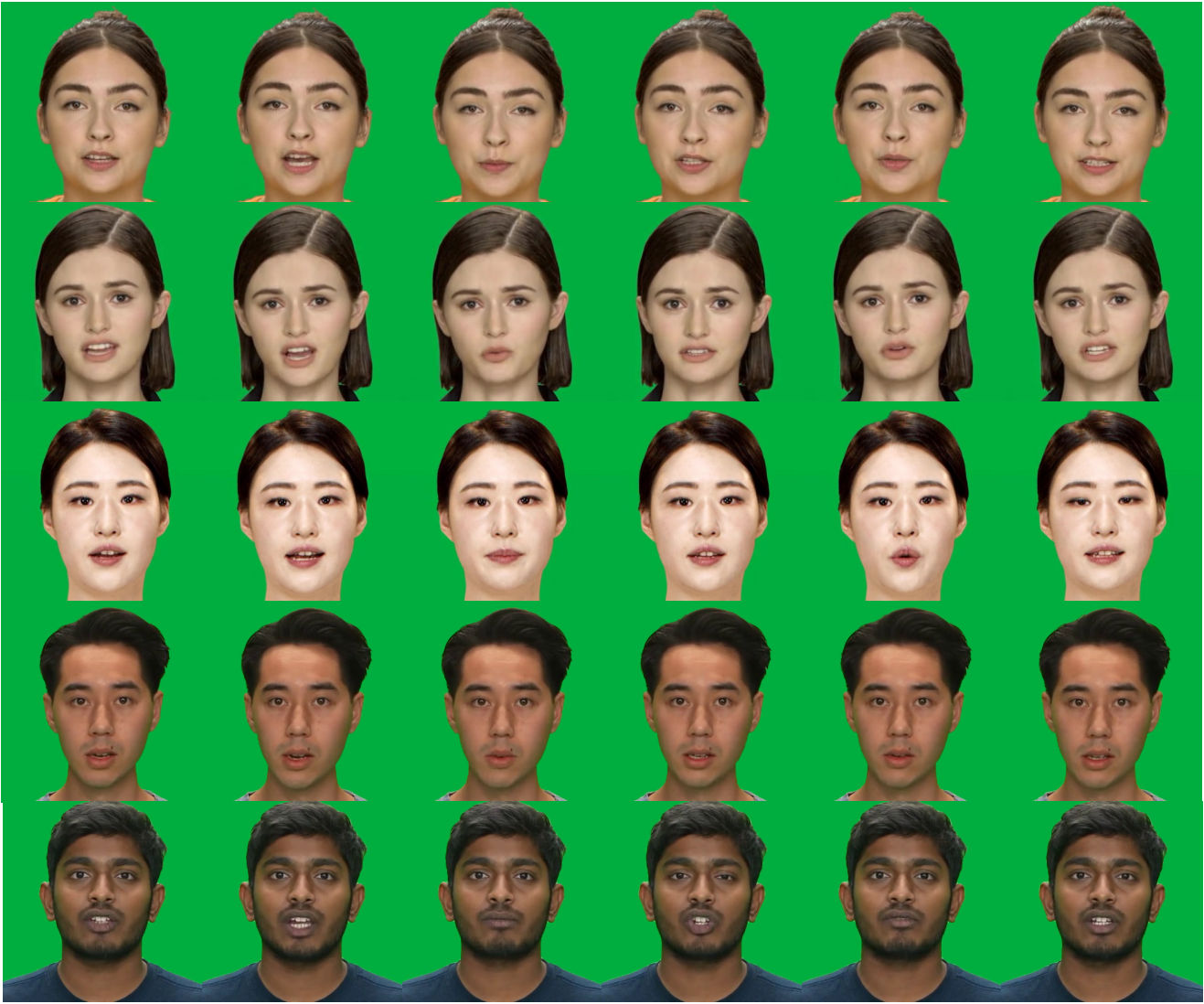}\caption{
		Results of our method on five different identities. 
		Zoom into the figure to better see lips and overall texture quality.
	}\label{fig:results_big}
\end{figure*}

{\small
\bibliographystyle{ieee_fullname}
\bibliography{main}
}

\end{document}